\documentclass[10pt,twocolumn,letterpaper]{article}

\usepackage{icb}
\usepackage{times}
\usepackage{epsfig}
\usepackage{graphicx}
\usepackage{amsmath}
\usepackage{amssymb}
\usepackage{amsmath}
\usepackage{subcaption}

\DeclareMathOperator*{\argmin}{arg\,min}

\usepackage[pagebackref=true,breaklinks=true,letterpaper=true,colorlinks,bookmarks=false]{hyperref}

\icbfinalcopy 

\ificbfinal\fi
\begin{document}

\title{Improving Fingerprint Pore Detection with a Small FCN}

\author{Gabriel Dahia \hspace{2cm} Maur\' icio Pamplona Segundo \\
Department of Computer Science, Federal University of Bahia \\
{\tt\small gdahia@gmail.com, mauriciops@ufba.br} \\
}

\maketitle
\thispagestyle{empty}

\begin{abstract}
  In this work, we investigate if previously proposed CNNs for fingerprint pore detection overestimate the number of required model parameters for this task.
  We show that this is indeed the case by proposing a fully convolutional neural network that has significantly fewer parameters.
  We evaluate this model using a rigorous and reproducible protocol, which was, prior to our work, not available to the community.
  Using our protocol, we show that the proposed model, when combined with post-processing, performs better than previous methods, albeit being much more efficient.
  All our code is available
  \ificbfinal
  at \url{https://github.com/gdahia/fingerprint-pore-detection}.
  \else
  in the Supplementary Materials.
  \fi
\end{abstract}

\section{Introduction}
\label{sec:introduction}
High-resolution fingerprint images allow the analysis of level 3 features~\cite{jain-pores-and-ridges}, with pores being a popular choice among researches~\cite{direct-pore, ridge-reconstruction, feature-guided}.
The high number of pores per fingerprint, for example, allows recognizing fingerprints even when using only partial images~\cite{ridge-reconstruction, td-sparse}.
In addition, combining them with traditional minutiae-based methods improves fingerprint recognition~\cite{direct-pore, feature-guided}.

All automatic pore-based fingerprint recognition methods require detecting pores first, creating a demand for accurate and efficient pore detection algorithms.
Recent works have tackled pore detection in high-resolution images with convolutional neural networks (CNN).
We believe this is a promising approach since results from these methods are a substantial improvement over previous ones~\cite{u-net, su-pores-deep}.

However, it is also our opinion that the current use of CNNs overestimates the number of model parameters required to accomplish accurate pore detection, leading to inefficient networks.
What is worse, the literature does not yet have a detailed protocol for evaluating pore detection.
As authors of previous works do not detail the evaluation protocols of their works enough to reproduce them and do not make their code publicly available, this makes comparing the performance of previous methods impossible.

To solve both problems, we first propose a simple fully convolutional neural network (FCN) architecture, with many fewer parameters than previous CNN methods.
To validate it, we devise and detail a protocol that uses the standard benchmark for high-resolution fingerprints, the Hong Kong Polytechnic University (PolyU-HRF) dataset~\cite{direct-pore}.
This protocol has the intention of standardizing pore detection evaluation, making it reproducible and directly comparable.
In order to make this goal easier to attain, we make all of our code publicly available.

To point discrepancies between the previously used validations and the proposed evaluation protocol, we also employ it to evaluate the two pore detection methods that report having the best results~\cite{ridge-reconstruction, su-pores-deep}.
In it, our FCN, combined with post-processing based on thresholding and non-maximum suppression (NMS), performs on par with the state-of-the-art, even with 200 times fewer parameters.

Our contributions with this work are:
\begin{itemize}
  \item we describe a simple FCN architecture and a post-processing step to enhance its results (Section~\ref{sec:pore-detection});

  \item we detail a reproducible evaluation protocol for pore detection using the PolyU-HRF dataset (Section~\ref{sec:experiments});

  \item we evaluate ours and two other methods with this new protocol (Section~\ref{sec:results}).
    We find evidence that our protocol is stricter than previous ones.
    Also, these experiments provide strong evidence that our method is the state-of-the-art for pore detection.

\end{itemize}

\section{Related work}
\label{sec:related-work}
Previously, methods approached pore detection with hand-crafted methods or heuristics.
Pore extraction with adaptive pore modeling~\cite{adaptive-pore-modeling} convolves the input fingerprint images with a pore model, acting as a matched filter, and determines that locations with strong response are pores.
The pore model is instantiated from a dynamic anisotropic pore model parametrized with local ridge orientation and pore scale.
The response map is post-processed by removing pores that are not located on ridges, thresholding and converting connected components of detections into single proposals to obtain the final detections.
We refer to this as the traditional post-processing approach for pore detection.

Adaptive pore modeling can be improved by considering the spatial distribution of pores~\cite{spatial-analysis}.
This approach computes the expected distance between adjacent pores as an estimate of the ridge width of non-overlapping 
blocks of the fingerprint image.
This is input to a classifier that is used to discard false detections made by the standard approach.

Dynamic pore filtering~\cite{ridge-reconstruction} consists of finding the nearest black (ridge) pixels above, below, to the right, and to the left of each white pixel.
Then, it counts the number of black-white transitions in the circumference delimited by these four points to decide whether the pixel is a pore or not.

CNNs have recently been proposed as alternatives to handcrafted pore detection.
Su \etal~\cite{su-pores-deep} train a 7 layer CNN to classify image patches as centered on pores or not.
Image patches are classified as centered on pores if they are centered within 3 pixels from a pore position in the ground truth.
With the trained CNN, inference is performed by classifying the image at every possible patch location.
To allow inferring for entire images, they propose to convert the last two fully connected layers into convolutional layers.
The memory cost of doing this, however, is substantial, as the first fully connected layer has 4096 units.

Wang \etal~\cite{u-net} use a U-Net to detect pores in fingerprint images.
The U-Net is trained to classify each spatial location in the image into one of three categories: pore centroid, ridge or background.
To detect pores, 20 patches are extracted from the input image and each is forwarded through the trained CNN.
This results in 3 probability maps per patch, one indicating the probability of pores, the other of ridges and the last of background region.
The ridge probability map for each patch is used to post-process the one for pores.
Afterward, the pore probability maps are binarized.
To combine the predictions for each patch, a boolean ``or'' operation is performed with all of them.

\section{FCN pore detection}
\label{sec:pore-detection}
To detect pores in fingerprint images, we train an FCN to determine which image patches are centered around a pore.
This model is trained using image patches, instead of entire images.
This is done to decorrelate examples in a batch and avoid overfitting because the datasets for this task have very few annotated images.
However, inference can be done by performing a single forward pass per image in the model.
We also propose using a post-processing method to enhance the FCN's outputs.
Without this, our model proposes multiple detections for a single pore.

\subsection{Model architecture}
\label{sec:model-architecture}
Table~\ref{table:model-architecture} shows our FCN architecture for pore detection.
All four layers use batch normalization~\cite{batchnorm}, valid padding, and stride equal to 1 in each dimension.
In the first three layers, the kernel size is ${3 \times 3}$ and the activations are ReLU; each is followed by ${3 \times 3}$ max pooling.
The number of filters is 32, 64 and 128 from first to third layer, respectively.
The output layer is preceded by dropout~\cite{dropout}, has kernel size of ${5 \times 5}$, a single filter, and sigmoid activation.

\begin{table}[h]
  \begin{center}
    \small
    \begin{tabular}{c|c|c|c}
      \textbf{\#} & \textbf{Layer} & \textbf{Size} & \textbf{Filters} \\ \hline
      1 & Conv + ReLU	& $3 \times 3$ & 32 \\
        & BatchNorm	& - & - \\
        & Max Pooling	& $3 \times 3$ & - \\
      2 & Conv + ReLU	& $3 \times 3$ & 64 \\
        & BatchNorm	& - & - \\
        & Max Pooling	& $3 \times 3$ & - \\
      3 & Conv + ReLU	& $3 \times 3$ & 128 \\
        & BatchNorm	& - & - \\
        & Max Pooling	& $3 \times 3$ & - \\
        & Dropout & - & - \\
      4 & Conv & $5 \times 5$ & 1 \\
        & BatchNorm	& - & - \\
        & Sigmoid	& - & - \\
    \end{tabular}
  \end{center}
  \vspace{-8pt}
  \caption{Pore detection FCN architecture.
  Every layer uses valid padding and unit strides.
  The architecture is adequate for detecting pores because the required information to decide if a region is centered on a pore fits entirely in the network's receptive field.}
  \label{table:model-architecture}
\end{table}

Using an FCN allows us to perform inference for images of arbitrary size, without requiring adaptations, like networks with fully connected layers do.
It also keeps the number of model parameters relatively low.

When applied to an ${M \times N}$ image, our network produces an ${(M - 16) \times (N - 16)}$ output that is a spatial map indicating the probability of each pixel being centered on a pore.
If the input image has dimensions ${17 \times 17}$, the output is a scalar, indicating the probability of the image being centered on a pore.
Thus, one can see it as a sliding window passing through overlapping image patches and outputting pore probabilities for each one of them.

Pore detection differs from general object detection, in which the sliding window approach we employ is no longer the state-of-the-art.
We argue that this architecture suits the pore detection task for two reasons.
First, region proposal in general object detection must consider object sizes that vary significantly since the objects to be detected can be as wide or as tall as the entire image.
This is not the case for pore detection.
Pores fit entirely in ${17 \times 17}$ patches since the area they occupy in 1200dpi images range from 5 to 40 pixels~\cite{su-pores-deep}.
Second, though providing global context 
significantly improves general object detection~\cite{yolo-9000, faster-rcnn}, it does not has the same effect for pores.
Most of the necessary context to determine if a patch is centered on a pore - the pixel intensities of ridges and valleys, the ridge orientation - is visible in the patch itself.
\subsection{Training problem formulation}
\label{sec:training}
The labeling for the training of our detection FCN follows the formulation of a previous work~\cite{su-pores-deep}.
The probability of an image patch $I_p$ of dimensions ${17 \times 17}$ having a pore is 1 if there is a pore within a bounding box of dimensions ${7 \times 7}$ centered in $I_p$'s center; the probability is 0, otherwise.
Training, then, is formulated as minimizing the cross-entropy between the model's predictions and the training data pore probability distribution.

While it is possible to train the CNN end-to-end for whole images, this is suboptimal when the number of training images is very small.
It is known that independent batch sampling is crucial for training neural nets~\cite{deeplearningbook} and for the effectiveness of batch normalization~\cite{batchrenorm}.

In our model architecture, using an entire image of dimensions ${M \times N}$ in a mini-batch is equivalent to using ${(M - 17) (N - 17)}$ highly correlated image patches.
The pixel intensities of the image, which can be roughly divided into two classes, ridge (darker pixels) and valley (lighter pixels), display little to no intraclass variation.
Also, overlapping and neighboring patches differ from one another only by small translations and a few pixels.

\begin{figure}[h]
  \begin{center}
  \includegraphics[width=0.35\textwidth]{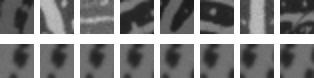}
  \end{center}
  \vspace{-8pt}
  \caption{Image patch correlation and patch sampling.
  The top row displays patches sampled randomly across images, while the bottom row displays patches sampled sequentially from a single image, which is equivalent to use entire images in mini-batches.
  Notice the variations in intensities of ridges and valleys across patches in the top row, and how ridge orientations vary among them.
  These variations are absent from sequentially sampled patches.
  }
  \label{fig:patch-sampling}
\end{figure}

We sidestep this problem by using patch sampling.
Patch sampling consists of sampling patches randomly from all images of the training set to perform gradient-based learning, instead of using entire images in mini-batches.
By doing this, the patches are as independent as possible.
Figure~\ref{fig:patch-sampling} illustrates the problem and the patch sampling approach.

\subsection{Post-processing}
\label{sec:post-processing}

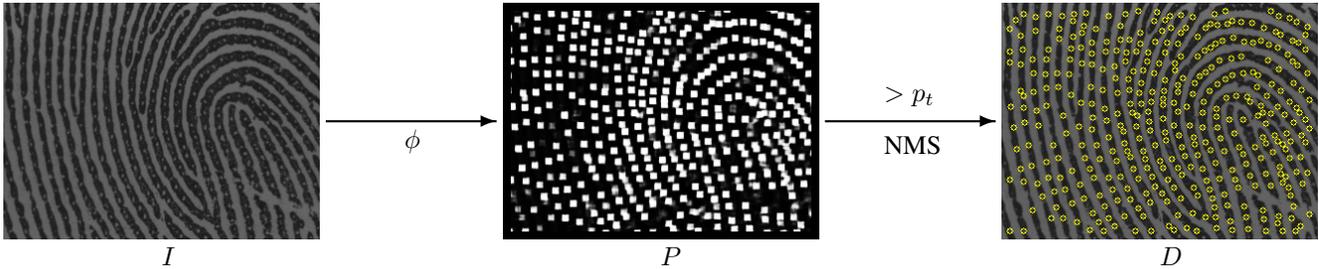
\begin{figure*}[t]
  \begin{center}
    \setlength{\unitlength}{\textwidth}
    \begin{picture}(1, 0.2)
      \thicklines
      \put(0, 0){\includegraphics[width=0.24\textwidth]{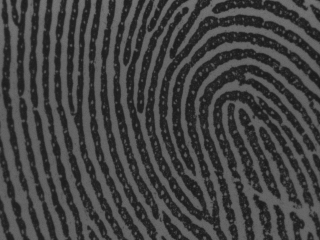}}
      \put(0.12, -0.02){$I$}

      \put(0.245, 0.09){\vector(1, 0){0.13}}
      \put(0.305, 0.07){$\phi$}

      \put(0.38, 0.00){\includegraphics[width=0.24\textwidth]{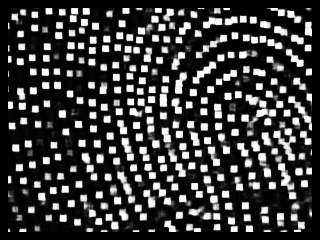}}
      \put(0.5, -0.02){$P$}

      \put(0.625, 0.09){\vector(1, 0){0.13}}
      \put(0.67, 0.105){$> p_t$}
      \put(0.67, 0.065){NMS}

      \put(0.76, 0){\includegraphics[width=0.24\textwidth]{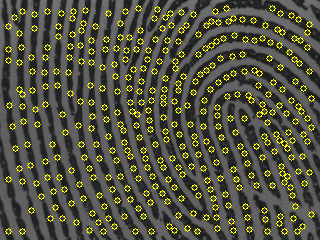}}
      \put(0.88, -0.02){$D$}

    \end{picture}
  \end{center}
  \vspace{-4pt}
  \caption{Pore detection pipeline.
  First, $P$ is obtained from the input image $I$ by zero-padding the output $\phi(I)$ of the detection CNN to input dimensions ${M \times N}$.
  Brighter pixels in $P$ indicate higher probability.
  $P$ is thresholded and the remaining points are converted to bounding boxes.
  We then merge multiple detections of the same pore with NMS and convert the bounding boxes back to coordinates to obtain $D$.
  Pore detections are circled in yellow.
  Best viewed in color.
  }
  \label{fig:post-processing}
\end{figure*}

Traditionally, models like ours propose every location in which the output indicates more than 50\% probability as a pore detection.
Visualizing these proposals showed us that there were multiple detections for each pore.
To address that problem, we adapted traditional object-detection post-processing techniques to the pore detection task.

Figure~\ref{fig:post-processing} displays the complete pore detection pipeline when combined with post-processing.
Let $I$ be the input image of dimensions ${M \times N}$ and ${P = \phi(I)}$ be the CNN's output, zero-padded to the input dimensions.
$P$ represents the probability of each spatial location of $I$ being centered on a pore.

We then construct the set $B$ of ${7 \times 7}$ bounding boxes around each spatial coordinate $(i, j)$ for which ${P(i, j)}$ is above a pre-determined threshold $p_t$.
The construction of $B$ is derived from how patches are labeled for training: probability 1 is assigned if a ${7 \times 7}$ bounding box around the center of the patch contains a pore.

The procedure outlined so far connects the bounding box detections of the same pore by the overlap in their areas.
Hence, we apply NMS to convert them into a single detection.
Converting the remaining bounding boxes to coordinates again results in the final detections, $D$.

\section{Experiments}
\label{sec:experiments}

We conduct our experiments in the default benchmark for pore detection, the PolyU-HRF dataset~\cite{direct-pore}.
Polyu-HRF is divided into three subsets: \textit{GroundTruth}, \textit{DBI}, and \textit{DBII}.
\textit{GroundTruth} consists of 30 1200dpi images annotated with coordinates for pores. 
\textit{DBI} and \textit{DBII} are for fingerprint recognition experiments and lack pore coordinate annotation.
Therefore, we use 
\textit{GroundTruth} 
for our experiments.

It is important to notice that there is no standard protocol to evaluate pore detection in \textit{GroundTruth}.
Because of this, authors follow their own protocols to validate their methods.
Furthermore, these protocols usually do not have the required level of detail to be reproduced.
Hence, we propose an evaluation protocol for pore detection in \textit{GroundTruth}, describe it in detail and provide code to reproduce it.
Our intention is that future works use our protocol, making reported results comparable.

Our evaluation protocol for pore detection in \textit{GroundTruth} is as follows.
Its 30 images should be split with the first 15 images forming the training set, the next 5 the validation set, and the last 10 the test set.

Evaluated methods should be compared by reporting true detection rates (TDR), false detection rates (FDR), and the resulting {F-score}.
TDR is computed as the recall of detections and FDR is computed as the false discovery rate of detections.
To compute these metrics, we further establish how to determine which detections are true detections and which are false detections.
Given a set of detections $D$ for an image and its corresponding ground truth $G$, first compute the distances of all pairs ${(d, g) \in D \times G}$.
Consider true detections only the $d \in D$ for which the distance $||d - g||_2$ for some $g \in G$ is both the minimum of the distances of $d$ and $g$, \ie:
\begin{equation}
  g = \argmin_{g' \in G} ||d - g'||_2
\end{equation}
and
\begin{equation}
  d = \argmin_{d' \in D} ||d' - g||_2.
\end{equation}
Every other detection is considered a false detection.

To allow a fair comparison between methods that use CNNs and the others, only pores that can be detected should be considered, \eg for our method and Su~\etal's~\cite{su-pores-deep}, pores in the first and last 8 rows and columns are not considered.

To validate our method, we conduct two experiments.
The first experiment is an ablation study of the post-processing technique.
We compare the results obtained when using the same FCN model 
with both post-processing as proposed and using the traditional approach~\cite{adaptive-pore-modeling}.
When using the FCN with traditional post-processing, a detection is proposed everywhere the output predicts more than 50\% probability of a pore.

The second experiment consists of evaluating, using the proposed protocol, the two best methods in terms of originally reported {F-score}.
This experiment allows not only the direct comparison of our method to previous ones but also an indirect comparison of our evaluation protocol to theirs.
If a method performs worse in the proposed protocol than in it did originally, this is evidence that our protocol is stricter.

\begin{figure*}[t]
  \begin{center}
    \begin{subfigure}{0.24\linewidth}
      \includegraphics[width=\textwidth]{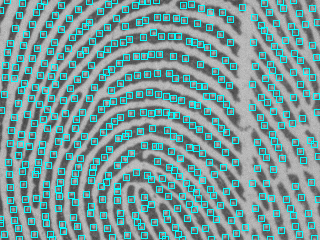}
      \caption{\label{fig:qualitative-ground-truth}Ground truth}
    \end{subfigure}
    \begin{subfigure}{0.24\linewidth}
      \includegraphics[width=\textwidth]{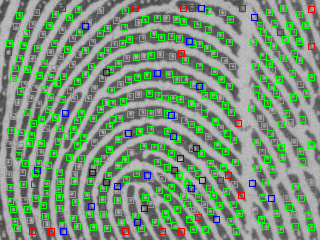}
      \caption{\label{fig:qualitative-proposed}Proposed}
    \end{subfigure}
    \begin{subfigure}{0.24\linewidth}
      \includegraphics[width=\textwidth]{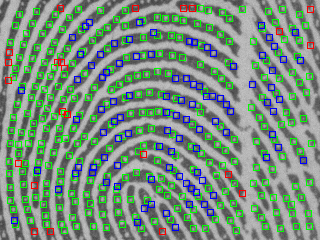}
      \caption{\label{fig:qualitative-su}Su \etal \cite{su-pores-deep}}
    \end{subfigure}
    \begin{subfigure}{0.24\linewidth}
      \includegraphics[width=\textwidth]{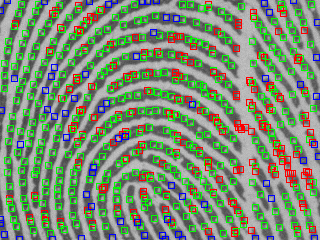}
      \ificbfinal
        \caption{\label{fig:qualitative-dpf}Pamplona II \& Lemes \cite{ridge-reconstruction}}
      \else
        \caption{\label{fig:qualitative-dpf}Segundo \& Lemes \cite{ridge-reconstruction}}
      \fi
    \end{subfigure}
  \end{center}
  \vspace{-8pt}
  \caption{Qualitative comparison of pore detection methods using the proposed evaluation protocol.
  The image is the last from the test set and was picked before the authors saw results for it.
  Green squares mark true positives (true detections), red squares mark false positives (false detections), and blue squares mark false negatives. 
  For the proposed method (\ref{fig:qualitative-proposed}) and Su~\etal's (\ref{fig:qualitative-su}), only pores that can be detected are considered.
  Best viewed in color.
  }
  \label{fig:qualitative}
\end{figure*}

\ificbfinal Pamplona \fi Segundo \& Lemes provided an implementation of their method~\cite{ridge-reconstruction}.
Since we were unable to obtain an implementation, the required detections or a trained model for Su~\etal's~\cite{su-pores-deep} method, we reimplemented it according to what is described in the paper.
However, we were unable to reproduce their results.
We attribute our failure to what we consider insufficient detailing of the method.
Specifically, there is neither a description of how to convert the proposals into detections nor how to determine which detections are true and which are false.
Since this conversion is not described, we assumed they used the traditional post-processing method~\cite{adaptive-pore-modeling}.
To improve transparency, we make our reimplementation of this method publicly available.

All the neural network models are optimized using Stochastic Gradient Descent (SGD) with early stopping.
Our code uses Tensorflow, NumPy, and OpenCV.
We make the code required to reproduce our experiments, alongside trained models, publicly available.

We manually tune the learning rate and its decay schedule, the batch size, the dropout rate and the amount of weight decay by measuring the {F-score} of the patches classified as pores using the training labels.
We set their values at $10^{-1}$, exponentially decayed by 0.96 every 2000 steps, 256, 0.2 and 0, respectively.
The post-processing parameters are obtained by performing a grid search, optimizing validation {F-score} with the trained model for $p_t$ and the NMS intersection threshold $i_t$.
The range of the search for $p_t$ is $\{0.1, 0.2, ..., 0.9\}$ and for $i_t$ is $\{0, 0.1, ..., 0.7\}$.
The chosen values were ${p_t = 0.6}$ and ${i_t = 0}$.
In our implementation of NMS, ${i_t = 0}$ corresponds to discarding bounding boxes that have any amount of intersection.

\section{Results}
\label{sec:results}

Our results for the post-processing ablation study are presented in Table~\ref{table:ablation}.
They reveal that using our method with the proposed post-processing is better than using the traditional one, according to the {F-score} metric.
The FDR for traditionally post-processed detections is 0.06\% lower in absolute than the proposed method, while its TDR is 16.85\% lower in absolute values.
This shows that the proposed post-processing technique merges detections of the same pore better than the traditional approach.

\begin{table}[h]
  \begin{center}
    \begin{tabular}{l|c|c|c}
      \textbf{Post-processing}      & \textbf{TDR}      & \textbf{FDR}      & \textbf{F-score} \\ \hline
      \textbf{Proposed}             & \textbf{91.95\%}  & 8.88\%            & \textbf{91.53\%} \\ \hline
      Traditional                   & 75.10\%           & \textbf{8.82\%}   & 82.36\% \\
    \end{tabular}
  \end{center}
  \vspace{-8pt}
  \caption{Post-processing ablation study.
  Comparison of pore detection results with the proposed method, controlling for the post-processing method.
  Rows are sorted in order of decreasing F-score and the best results per column are in boldface.
  }
  \label{table:ablation}
\end{table}

\begin{table}[t]
  \begin{center}
    \begin{tabular}{l|c|c|c}
      \textbf{Method}                               & \textbf{TDR}      & \textbf{FDR}      & \textbf{F-score} \\ \hline
      \textbf{Proposed}                             & \textbf{91.95\%}  & \textbf{8.88\%}   & \textbf{91.53\%} \\ \hline
      Su \etal~\cite{su-pores-deep}                 & 70.77\%           & 11.58\%           & 78.58\% \\ \hline
      \ificbfinal
        Pamplona Segundo \&                         & 89.31\%           & 38.02\%           & 73.17\% \\
        Lemes~\cite{ridge-reconstruction}           &                   &                   & \\ 
      \else
        Segundo \& Lemes~\cite{ridge-reconstruction}& 89.31\%           & 38.02\%           & 73.17\% \\
      \fi
    \end{tabular}
  \end{center}
  \vspace{-8pt}
  \caption{Comparison of pore detection results using our evaluation protocol.
  The drop in performance for the previous two best pore detection methods indicate that our evaluation protocol is stricter than previous ones.
  Also notice that our method is the state-of-the-art for all three metrics in this protocol.
  Rows are sorted in order of decreasing {F-score} and the best results per column are in boldface.}
  \label{table:evaluated}
\end{table}

Figure~\ref{fig:qualitative} presents qualitative results and Table~\ref{table:evaluated} shows metrics for our second experiment.
Table~\ref{table:reported} provides reported results for comparison.
The depicted fingerprint was taken from the test set and it was not seen before it was chosen.
Nevertheless, we consider it representative of the quantitative results of our second experiment.

First, we remark that while \ificbfinal Pamplona \fi Segundo \& Lemes' method~\cite{ridge-reconstruction} finds many true detections (green squares in the figure), it also proposes many detections which are not pores (red squares).
This result is observed not only in our experiments but in this work's original validation as well.
The statistics obtained for this method are perfectly comparable to ours, because the authors provided us their implementation.
These results show that the proposed method is an improvement over \ificbfinal Pamplona \fi Segundo \& Lemes' and that the employed evaluation protocol is stricter.

As for Su~\etal's method~\cite{su-pores-deep}, one can see in Figure~\ref{fig:qualitative} that it does not make many false detections.
Our reimplementation, however, fails to detect a significant number of pores (blue squares in the figure), which impacts its TDR in our experiments.
Despite reporting high TDR (88.6\%), the qualitative results shown in Su~\etal's work display many undetected pores, which is consistent with our findings.

\begin{table}[t]
  \begin{center}
    \begin{tabular}{l|c|c|c}
      \textbf{Method}                                & \textbf{TDR}      & \textbf{FDR}   & \textbf{F-score} \\ \hline
      Su \etal~\cite{su-pores-deep}                  & 88.6\%            & \textbf{0.4\%} & \textbf{93.78\%} \\ \hline
      \textbf{Proposed}                              & \textbf{91.95\%}  & 8.88\%         & 91.53\% \\ \hline
      \ificbfinal
        Pamplona Segundo \&                          & 90.80\%           & 11.10\%        & 89.84\% \\
        Lemes~\cite{ridge-reconstruction}            &                   &                & \\ \hline
      \else
        Segundo \& Lemes~\cite{ridge-reconstruction} & 90.80\%           & 11.10\%        & 89.84\% \\ \hline
      \fi
      Teixeira \& Leite~\cite{spatial-analysis}      & 86.10\%           & 8.60\%         & 88.67\% \\ \hline
      Wang \etal~\cite{u-net}                        & 83.65\%           & 13.89\%        & 85.88\% \\ \hline
      Zhao \etal~\cite{adaptive-pore-modeling}      & 84.80\%           & 17.60\%        & 83.58\% \\
    \end{tabular}
  \end{center}
  \vspace{-8pt}
  \caption{Reported pore detection results.
  These results are not comparable because each work adopts its own validation method.
  Rows are sorted in order of decreasing {F-score} and the best results per column are in boldface.}
  \label{table:reported}
\end{table}

The decrease in Su~\etal's reported performance is more evidence that the proposed protocol is stricter than previous validation methods.
However, due to ours being a reimplementation and the lack of explanation in how detections are post-processed in that paper, we cannot guarantee this.

The proposed method not only outperforms our reimplementation of Su~\etal's method~\cite{su-pores-deep} in the proposed evaluation protocol, it does so even when our method is traditionally post-processed.
We consider this comparison fair because the optimization method, the effort put into optimizing the models, and the dataset split were the same for both methods.
Moreover, we make the code for our method and the reimplementation of theirs publicly available.

We believe these arguments and the provided quantitative results are strong evidence to support our claim that the proposed evaluation protocol is stricter than previous validation methods.
Accepting these evidence is enough to make the proposed method the state-of-the-art for the pore detection task.
We further remark that our model has 96,548 parameters, in contrast to 21,211,074~\cite{su-pores-deep} or 1,696,512~\cite{u-net} parameters of the previous neural network methods.
We believe this improvement is due to the use of a more adequate number of parameters for the task and the effectiveness of our post-processing method.

Analyzing the visual results of the proposed method in Figure~\ref{fig:qualitative} provides some insight into what causes it to make a mistake.
Most of the false positives of our method fall in two cases.
The first is caused by pores that are visible in the region for which proposals are possible but whose centers are in the non-visible borders.
This causes these pores to be discarded in our protocol.
The same can be observed for our reimplementation of Su~\etal's method~\cite{su-pores-deep}.
The second case is for image regions in which it is difficult to distinguish if it is not, in fact, a pore.
These can be, other than pores, lighter ridge regions, scars or bifurcations.

\section{Conclusion}

In this work, we trained an FCN to detect pores in high-resolution fingerprint images.
It is our opinion that, as the writing of this paper, it is impossible to perfectly compare all the pore detection methods.
This is because there is no default evaluation protocol in the standard pore detection benchmark, PolyU-HRF.
In our attempt to establish that the proposed method is the state-of-the-art, we conducted a rigorous and extensive empirical evaluation.
It includes an ablation study of our post-processing method, perfectly comparable results for one of the previous methods and our best efforts in reimplementing another.
The evidence we gather point to our method indeed being the state-of-the-art.

Our second contribution, if effective, can solve the problem of non-comparable results.
We propose a detailed and reproducible evaluation protocol, the first for this task, incredibly.
Not only this, but we make the code for reproducing all of our experiments available.

We see potential of improving pore detection using semi-supervised learning techniques with the unlabeled training data in PolyU-HRF's \textit{DBI} subset, because it has 14 times the images in the training split we propose.
Previous work on semi-supervised learning indicates this is enough to offer substantial improvement~\cite{semi-supervised}.

The main limitation of our method is its possible susceptibility to domain variations~\cite{domain-gan}.
Since PolyU-HRF's data is collected using a single sensor, it is possible that our pore detector fails to generalize to images collected with other sensors.
The difference in fingerprint images collected with different sensors is significant, as can be seen when comparing images from PolyU-HRF to those from other datasets~\cite{zhang-pattern-rec}.
If this is truly an issue, collecting more data and using domain adaptation might be necessary~\cite{domain-gan}.

\ificbfinal
\section*{Acknowledgments}
This work was funded by Universidade Federal da Bahia (UFBA) and Akyiama Solu\c c\~oes Tecnol\'ogicas.
The Titan Xp we used was donated by the NVIDIA Corporation.
\fi

{\small
\bibliographystyle{ieee}
\bibliography{main}
}

\end{document}